\documentclass[10pt]{article}

\usepackage{amsmath,amssymb,bm,bbm,graphicx,caption,adjustbox}
\usepackage{subcaption,mathtools,multirow}
\usepackage[linesnumbered,ruled]{algorithm2e}
\usepackage{color}
\usepackage{adjustbox}
\usepackage{array,ragged2e}
\usepackage{authblk}

\usepackage{float}  
\usepackage{hyperref}

\newlength\mylen
\newcommand\myinput[1]{
  \settowidth\mylen{\KwIn{}}
  \setlength\hangindent{\mylen}
  \hspace*{\mylen}#1\\}
\let\oldnl\nl
\newcommand{\nonl}{\renewcommand{\nl}{\let\nl\oldnl}}

\DeclareMathOperator{\argmax}{argmax}

\newcommand{\newterm}[1]{\textit{#1}}

\long\def\comment#1{}

\title{Distributed Entity Disambiguation with Per-Mention Learning}
\author[1]{Tiep Mai\thanks{Now at IBM, Ireland (stmai@ie.ibm.com)}}
\author[2]{Bichen Shi}
\author[1]{Patrick K. Nicholson}
\author[1]{Deepak Ajwani}
\author[1]{Alessandra Sala}
\affil[1]{Bell Labs, Nokia,
  \texttt{firstname.lastname@nokia.com}}
\affil[2]{University College Dublin, Ireland,
  \texttt{bichen.shi@insight-centre.org}}

\date{}

\begin{document}

\maketitle


\begin{abstract}

Entity disambiguation, or mapping a phrase to its canonical representation in a knowledge base, is a fundamental step in many natural language processing applications. 
Existing techniques based on global ranking models fail to capture the individual peculiarities of the words and hence, either struggle to meet the accuracy requirements of many real-world applications or they are too complex to satisfy real-time constraints of applications.

In this paper, we propose a new disambiguation system that learns specialized features and models for disambiguating each ambiguous phrase in the English language. 
To train and validate the hundreds of thousands of learning models for this purpose, we use a Wikipedia hyperlink dataset with more than 170 million labelled annotations. 
We provide an extensive experimental evaluation to show that the accuracy of our approach compares favourably with respect to many state-of-the-art disambiguation systems. 
The training required for our approach can be easily distributed over a cluster. Furthermore, updating our system for new entities or calibrating it for special ones is a computationally fast process, that does not affect the disambiguation of the other entities.

\end{abstract}

\setlength{\tabcolsep}{.15em}

\section{Introduction}
\label{sec:introduction}

Many fundamental problems in natural language processing, such as text
understanding, automatic summarization, semantic search, machine
translation and linking information from heterogeneous sources, rely
on entity disambiguation \cite{ref:Ferrucci2012,ref:Suchanek2013}. 
The goal of entity disambiguation and more generally,
word-sense disambiguation is to map potentially ambiguous words and phrases in
the text to their canonical representation in an external knowledge
base (e.g., Wikipedia, Freebase entries). This involves resolving the
word ambiguities inherent to natural language, such as homonymy
(phrases with multiple meanings)  and synonymy
(different phrases with similar meanings),  thereby, 
revealing the underlying semantics of the text.

\textbf{Challenges:} This problem has been well-studied for well over a decade and has seen
significant advances. However, existing disambiguation approaches
still struggle to achieve the required \textbf{accuracy-time trade-off} for supporting
real-world applications, particularly those that involve streaming text such as tweets,
chats, emails, blogs and news articles. 

A major reason behind the \textbf{accuracy} limitations of the existing
approaches is that they rely on a single global ranking model
(unsupervised or supervised) to map all entities. These approaches 
fail to capture the subtle nuances of individual
words and phrases in the language. Even if dictionaries
consider two words to be synonyms, the words often come from different
origins, evoke different emotional images, differ in their general
popularity and usage by demographic groups as well as how they relate to the local
culture. 
Hence, even synonymous
words can have very different probability distribution of being mapped
to different 
nodes in the knowledge base. Global ranking
models fail to capture the individuality of words and are, thus, more
prone to mistakes in entity disambiguation.

In terms of \textbf{running time}, there are three aspects:
\begin{enumerate}
\item Training time: Ideally, it
should be possible to break the training into a large number of small tasks for
distributed scaling. Unfortunately, disambiguation approaches based on a global
ranking model are either not amenable to parallelism at all or
non-trivial to parallelize efficiently.
\item Query time: The real-time requirements of many 
applications constrain the extracted features to be light-weight and
limits the complexity of the approaches. For instance,  many joint
disambiguation approaches are infeasible within real-time constraints.
\item Update time: Many applications require disambiguation systems to be
continually updated as new words and phrases gain relevance (e.g.,
``Ebola crisis'', ``Panama papers'', ``Migrant crisis'') and additional training
data becomes available. Ideally, extending a system for new phrases or
calibrating it for special ones should be local operations, i.e.,
they should not affect the previous learning of other word phrases and
they should be fast. Existing approaches based on global models
(supervised or unsupervised) often require the computationally
expensive operation of rebuilding the global model from scratch, every time there is an update.
\end{enumerate}

\noindent \textbf{Our Approach:} In this paper, we propose a novel approach to address all of these
issues in word-sense disambiguation. Our approach aims at learning
the individual pecularities of entities (words and phrases) in the
English language and learns a specialized classifier for each ambiguous
phrase. This allows us to find and leverage features that best
differentiate the different meanings of each entity.

To train the hundreds of thousands of classifiers for this
purpose, we use the publicly available Wikipedia hyperlink
dataset. This dataset contains more than 110 million annotations and
we extend it by another 63 million annotations. Since the training of these
classifiers is independent of each other, our approach can be easily
parallelized and we use a distributed Spark cluster for this
purpose. The features used in these classifiers are mostly based on
text overlap and are, therefore, light-weight enough for its usage in real-time
systems. Updating our system for new entities simply requires to learn
the models for those entities - it is fast and does not affect the
other classifiers. Also, our approach is more robust in the presence
of noisy data. Furthermore, unlike the increasingly popular deep
learning architectures, our approach is interpretable: it is easy to
understand why our models chose a particular mapping for an entity. We
provide an extensive experimental evaluation to show that the efficacy
and accuracy of our approach compares favourably with respect to many
state-of-the-art disambiguation systems.

\textbf{Outline:} The rest of the paper is organized as follows. 
Section~\ref{sec:rel} presents related disambiguation techniques.
Section \ref{sec:data} gives 
a brief overview of the Wikipedia hyperlink data used in the training
of our disambiguation system and describe a method to extend this annotation data. 
In Section \ref{sec:disambiguation}, we present the details of our
novel disambiguation approach. 
Section \ref{sec:ex_setup} describes the experiment set-up, comparison metrics, 
and also a method to scramble data sets for validation
of different scenarios. 
Experiment results with different training settings
on a part of Wikipedia data
are in Section \ref{sec:analysis01}. 
Then, Section \ref{sec:pruning_pr} shows our pruning method and results. 
Comparisons with Dbpedia Spotlight, TAGME and
benchmarking results by the standard framework GERBIL \footnote{http://aksw.org/Projects/GERBIL.html}
are given in Section \ref{sec:spotlight}. 
Experiment results of full Wikipedia data and per-phrase accuracy
are given in Section \ref{sec:full_exp}.
Finally, we conclude this paper with some points for future works
in Section \ref{sec:conclusion}.

\section{Related Work}
\label{sec:rel}

There is a substantial body of work focussing on the task of disambiguating
entities to Wikipedia entries. The existing techniques can be roughly categorized 
into unsupervised approaches that are mostly graph-based and
supervised approaches that learn a global ranking model for
disambiguating all entities.

\textbf{Graph-based unsupervised approaches:} In these approaches, a
weighted graph is generally constructed
with two types of nodes: phrases (mention) from the text and the candidate
entries (senses) for that phrase. For the mention-sense edges, the
weights represent the likelihood of the sense for the mention in
the text context such as those given by the overlapping metrics between the mention context and 
the sense context (Wikipedia article body). For the sense-sense edges,
the weights capture their relatedness,
e.g. the similarity between two Wikipedia articles 
in terms of categories, in-links, out-links. A scoring function 
is designed and then optimized on the target document
so that a single sense is associated with one mention. 
Hoffart~\cite{ref:Hoffart2013} solved the optimization
by a dense subgraph algorithm; on the other hand, 
Han et al.~\cite{ref:Han2011B} as well as Guo and
Barbosa~\cite{GuoB14} used random walk on the graph and chose the 
candidate senses by the final state probability. Hulpus et
al.~\cite{HulpusPH15} explored some path-based metrics for joint
disambiguation. The AGDISTIS approach of Usbeck et
al.~\cite{AGDISTIS2014} extracts a subgraph of DBpedia graph
containing all the candidate senses and uses a centrality measure
based on HITS algorithm on the extracted subgraph to score the
senses. It then selects the sense with the highest authority score for each
entity. Moro et al.~\cite{ref:Moro2014} leveraged BabelNet
\footnote{http://babelnet.org}  to constructed a semantic graph 
in a different manner, where each node is a 
combination of mention and candidate sense. 
Thereafter, a densest subgraph is extracted and the senses with
maximum scores are selected. Ganea et al.~\cite{PBoH2016}
consider a probabilistic graphical model (PBoH) that addresses collective
entity disambiguation through the loopy belief propagation. 

Since these graph-based solutions are mostly unsupervised, there is no 
parameter estimation or training during the design 
of the scoring function to guarantee the compatibility 
between the proposed scoring function and the observed errors
in any trained data
\cite{ref:Han2011B,ref:Hoffart2013,ref:Piccinno2014}. 
Some disambiguation systems do apply a training phase on the final
scoring function (e.g., TAGME by Ferragina and
Scaiella~\cite{ref:Ferragina2010B}), but even here, the learning is
done with a global binary ranking classifier. 
An alternative system by Kulkarni et al.~\cite{ref:Kulkarni2009} uses
a statistical graphical model where the unknown senses are treated as latent variables 
of a Markov random field.
In this work, the relevance between mentions
and senses is modelled by a node potential and 
trained with max-margin method. The trained potential
is combined with a non-trained clique potential,
representing the sense-sense relatedness, to 
form the final scoring function. However, maximizing this scoring function
is NP-hard and computationally intensive \cite{ref:Ferragina2010B}.

\textbf{Supervised global ranking models:} 
On the other hand, non-graph-based solutions 
\cite{ref:Daiber2013,ref:Han2011A,ref:McNamee2010,ref:Meij2012,ref:Milne2008,ref:Olieman2014}
are mostly supervised in the linking phase. 
Milne and Witten~\cite{ref:Milne2008} assumed that there exists unambiguous mentions 
associated with a single sense, and evaluated
the relatedness between candidate senses and unambiguous mentions (senses).
Then, a global ranking classifier is applied on 
the relatedness and commonness features. 
Not relying on 
the assumption of existing unambiguous mentions,  
Cucerzan~\cite{ref:Cucerzan2007} constructed document attribute vector as
an attribute aggregation of all candidate senses and
used scalar product to measure different 
similarity metrics between 
document and candidate senses. While the original method
selected the best candidate by an unsupervised scoring function,
it was later modified to use a global logistic regression model
\cite{ref:Cucerzan2014}.

In \cite{ref:Han2011A}, Han et al. proposed 
a generative probabilistic model, using the frequency of mentions
and context words given a candidate sense, as 
independent generative features; this statistical model is also 
the core module of the public disambiguation service
Dbpedia Spotlight \cite{ref:Daiber2013}. 
Then, Olieman et al. proposed various adjustments 
(calibrating parameters, preprocessing text input, 
merging normal and capitalized results) to adapt Spotlight
to both short and long texts \cite{ref:Olieman2014};
Olieman et al. also used a global binary classifier with 
several similarity metrics to prune off uncertain 
Spotlight results.

In contrast to these approaches that learn a global ranking model for
disambiguation, our approach learns specialized features and model for
each mention. Furthermore, since the disambiguation learning is per-mention and
the number of candidate senses is fixed per-mention, 
proper multi-class statistical model can be used instead
of binary ranking classifier, and coherent predicted probability-across-class can be evaluated
and used for subsequent analysis.

\textbf{Per-mention disambiguation:} 
In terms of per-mention disambiguation learning 
on the Wikipedia knowledge base, 
Qureshi et al.'s~\cite{ref:Qureshi2014} method is the most similar 
to our proposed method. However, as their method 
only uses Wikipedia links and categories for feature design
and is trained with a small Twitter annotation dataset (60 mentions), 
it  does not fully leverage the rich and big Wikipedia annotation data 
to obtain highly accurate per-mention trained models. 
Also, while our feature extraction procedure is light and
tuned to contrast different candidate senses per mention, 
their method extracts related categories,
sub-categories and articles up to two depth level for each 
candidate sense, and requires pairwise relatedness scores between 
candidate sense and context senses. All these high cost features are 
computed on-the-fly due to the dependency on the context, 
potentially slowing down the disambiguation process.

\textbf{Pruner:} 
After the linking phase, most disambiguation
systems use either binary classifiers or score thresholding
to prune off uncertain annotations, trading off between 
precision and recall depending on the scenario
\cite{ref:Ferragina2010A,ref:Han2011A,ref:Milne2008,ref:Piccinno2014,ref:Olieman2014}.
As the recall values of our disambiguation are quite high,
we focus on increasing precision to guarantee non-noisy disambiguation outputs
for subsequent text analysis and applications. 
Both binary classifiers and thresholding are tested, as in previous methods. 
However, unlike these methods, our pruning is performed 
at per-mention and even per-candidate levels.

\section{Annotation Data and Disambiguation Problem}
\label{sec:data}

For the notation purposes, from this section, 
the first encounters of new terms are marked
by italic font, with their definitions provided in the corresponding places.

\subsection{Data preprocessing}
\label{subsec:data_preprocessing}

In this work, we use the publicly available Wikipedia dump 2015-07-29. 
This raw data is first processed by the script 
WikiExtractor\footnote{\url{http://medialab.di.unipi.it/wiki/Wikipedia_Extractor}}, 
which removes XML metadata, templates, lists, unwanted namespaces, etc. 

From the output data above, information for 
Wikipedia \newterm{entities} (\newterm{articles}) $e$ is extracted, including \newterm{article ids} ($e_{id}=e$), 
\newterm{article titles} ($e_{title}$), and \newterm{text bodies} ($e_{body}$).

In the text bodies of Wikipedia entities, there are hyperlink texts, linking text phrases 
to other Wikipedia entities. So, for the notation purposes, 
hyperlink texts are called \newterm{annotations}; their associated text phrases 
and Wikipedia entities are called \newterm{mentions} and \newterm{senses} accordingly
\footnote{Note that the term entity refers to a standalone Wikipedia article
while the term sense refers to a Wikipedia article linked by a mention.}.

Such annotations $a$, 
linking from mentions $m$ to Wikipedia senses $e$, 
contained in some Wikipedia articles $e^{\prime}$ 
are also extracted,
with information of particular annotations such as 
the \newterm{containing article ids} ($a_{cid}=e^{\prime}_{id}=e^{\prime}$), 
the mentions ($a_{m}=m$), the \newterm{destination senses} ($a_{e}=e_{id}=e$) 
and the short texts around the annotations (\newterm{annotation contexts} $a_{context}$)
\footnote{Redirections have been resolved by tracing from 
immediate annotation destinations
to the final annotation destinations $a_{e}$.}. 
For the annotation contexts, a number of sentences are extracted from both 
sides of the annotation so that the number of words of combined
sentences of each side exceeds a predefined \newterm{context window value} $n_{context}=50$
\footnote{Stopwords are counted in \newterm{annotation contexts}}.

During this process, text elements such as text bodies $e_{body}$, mentions $a_{m}$,
and annotation contexts $a_{context}$ are lemmatized using the python package nltk for the
purpose of grouping different forms of the same term.

\textbf{Disambiguation problem:} The extracted labelled annotations
are grouped by their mentions. Then, for a single unique mention $m$ such as ``Java'', 
we obtain the list of unique associated senses from the annotation group of mention $m$,
e.g. ``Java (programming language)'', ``Java coffee'', ``Java Sea'', etc,
and make them the \newterm{candidate senses} $e^{(i)}$ for mention $m$ with ($i=1, \ldots, |e|$
and $|e|$ is the number of candidate senses for mention $m$) 
\footnote{We drop index $m$ for $e^{(i)}$ and $|e|$ for simplicity.}.
In the \newterm{disambiguation problem}, given a new unlabelled annotation $a^{\prime}$ with 
its mention $a^{\prime}_{m}=m$ and context $a^{\prime}_{context}$, one wants to find 
correct sense $a^{\prime}_{e}=e^{\star}$ among all candidate senses $e^{(i)}$.

\subsection{Data extension}
\label{subsec:data_extension}

As data quantity is the key point in big data analysis in improving 
the performance of any learning algorithm, we want to enrich 
the annotation data as much as possible for the disambiguation problem.
Noticing that most Wikipedia annotations are only applied for 
the first occurrence of text phrases and article self-links are 
not available, we extend the annotation data with the following
extension procedure. 

First, unique annotation pairs $(a_{m}, a_{e})$ are extracted 
from each Wikipedia article $e^{\prime}$. 
If there is more than one annotation candidate article 
for a unique lemmatized mention $a_{m}$, the candidate article $e$ with 
the highest text overlapping between its title $e_{title}$
and annotation mention $a_m$ is selected. 
We also construct a unique self-annotation pair
$(a_{m} = e^{\prime}_{title}, a_{e} = e^{\prime})$ for each Wikipedia article $e^{\prime}$. 
This unique self-annotation pair takes precedence and overwrites any other pair of the
same mention in $e^{\prime}$.

Next, the body of each Wikipedia article is scanned, where 
lemmatized unlinked text phrases that exactly match a 
mention $a_{m}$ from the set of unique annotation pairs are gathered 
and paired with the corresponding senses $a_{e}$.
In this extended annotation dataset, original annotations are marked with 
extraction flags $a_{ef}=0$ while flags
$1$ and $2$ are for annotations extended from original pairs and Wikipedia
article titles, correspondingly. 

There is one general case to be noted with the above procedure. 
For example,
in the article $e^{(a)}_{title}=$ ``Biomedical engineering'', 
there is an original annotation
from $a^{(i)}_{m}=\text{``engineering''}$ to a more general article:
$a^{(i)}_{e}=e^{(b)}$ with 
$e^{(b)}_{title}=\text{``Engineering''}$. 
Any unlinked text phrase ``engineering'' in that article is 
annotated to $a^{(j)}_{e}=e^{(b)}$ with $a^{(j)}_{ef}=1$, 
which may not be true.
So, to deal with this issue, we remove the above annotations $a^{(j)}$ 
of flag $a^{(j)}_{ef}=1$ in this case.

Following this procedure, we are able to get around $63$ million additional annotations.
The union of original and extended annotation datasets
will be referred to as dataset $\mathcal{A}$ in the remainder of the paper.
As Wikipedia is a highly coherent and formal dataset, manually edited by many
people, we find the extended annotations to be of good quality.

The summary statistics of this data are given in Table \ref{tab_data:wikidatastat};
$\#\text{mentions}_{1}$ is the 
number of mentions that have exactly one candidate sense; 
$\#\text{mentions}_{>1}$ is for mentions with more than one
candidate senses.
In this paper, we target the disambiguation of $\text{mentions}_{>1}$.

\begin{figure}[h!tb]
\centering
\begin{subfigure}{0.48\textwidth}
    \centering
    \includegraphics[scale=0.65]{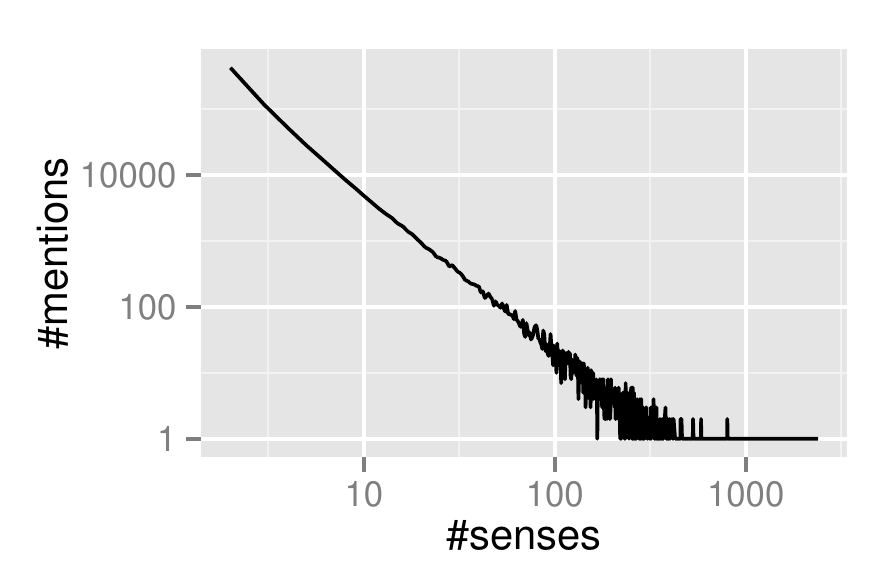}
    \caption{$\#\text{mentions}$ vs $\#\text{senses}$}
    \label{fig_data:sense_sf}
\end{subfigure}
\begin{subfigure}{0.48\textwidth}
    \centering
    \includegraphics[scale=0.65]{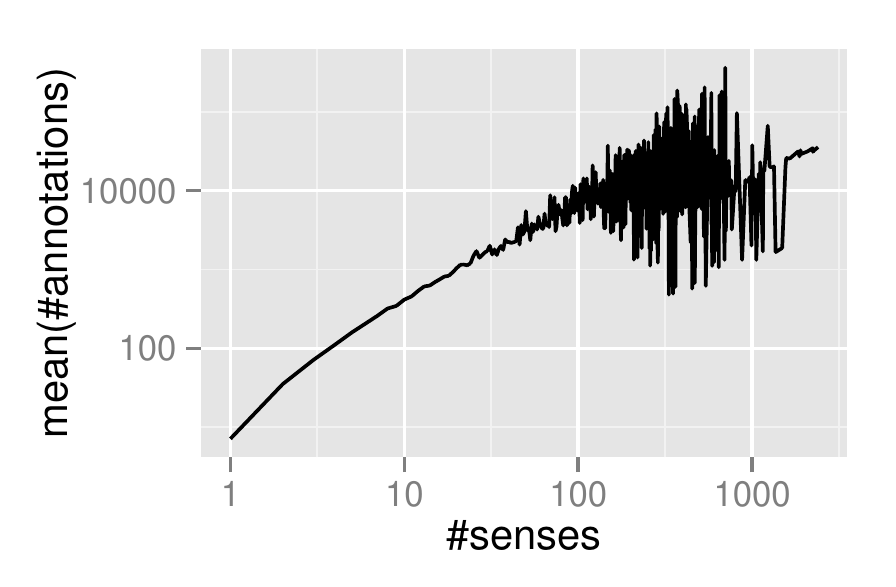}
    \caption{$\text{mean(\#annotations)}$ vs $\#\text{senses}$}
    \label{fig_data:sense_annotation}
\end{subfigure}
\caption{Summary statistics per sense}
\label{fig_data:sense_sf_annotation}
\end{figure}

Figure \ref{fig_data:sense_sf} shows the number of mentions
of a particular number of candidate senses in log-log scale. 
While most mentions have only a few candidates, some of them
such as ``France'', ``2010'' can have more than $600$ candidate senses 
as they are used to refer to numerous specific entities and 
events in a year or a country, such as ``French cinema'', ``Rugby league in France'',
``2010 America's Cup'', ``2010 in Ireland'', etc.
Figure \ref{fig_data:sense_annotation}
shows the averaged number of annotations of mentions of 
a particular number of senses. This number is proportional to the
training and validation size of the disambiguation learning per
mention, discussed in later sections. 

\begin{table}[t]
\caption{Wikipedia summary statistics}
\centering
\begin{tabular}{|l|r|}
\hline
$\#\text{main entities}$ & 4883834\\
\hline
$\#\text{redirection entities}$ & 6860485\\
\hline
$\#\text{mentions}_{1}$ & 8232340\\
\hline
$\#\text{mentions}_{>1}$ & 710044\\
\hline
$\#\text{original annotations}$ & 110003963\\
\hline
$\#\text{extended annotations}$ & 63375236\\
\hline
$\text{average context length}$ & 121.55\\
\hline
\end{tabular}
\label{tab_data:wikidatastat}
\end{table}

\section{Disambiguation Method}
\label{sec:disambiguation}

With Wikipedia annotation data extracted
and extended in the previous section, this section
discusses the learning method and feature extraction 
for the disambiguation problem.

As briefly mentioned in Section \ref{sec:introduction},
this paper uses a big data approach with 
supervised discriminative machine learning models
for the disambiguation problem.
Unlike methods 
where a single global supervised model is learned for all mentions \cite{ref:Han2011A, ref:Olieman2014} 
or one global ranking model is applied on a varying number
of candidate senses per encounter of mentions, 
\cite{ref:Ferragina2010A, ref:Ferragina2010B, ref:Milne2008}, 
our disambiguation system learns
and works on the basis of per lemmatized mention. 
Hence, annotations with the same lemmatized mention are grouped 
together for disambiguation learning. By doing so, the
problem of disambiguating a particular mention becomes 
a formal problem of multi-class classification,
which is, in our opinion, more statistically consistent and proper
than a ranking binary classifier in selecting 
a candidate sense.
\footnote{NIL-candidate-sense is not added at this stage 
but will be addressed by the pruner in the precision-recall
analysis}.

From the machine learning and statistical inference perspective, 
this per-mention learning approach is like segmenting varying big data into small data bins of 
similar properties and then applying the learning on each data bin
independently.
Within each data bin (or mention), a model and features
can be designed specifically for that particular mention.
As multiple specialized predictors perform better than 
a global multi-purpose predictor, the performance of the 
disambiguation is improving.

\subsection{Feature extraction}
\label{subsec:feature_extraction}

Most recent disambiguation works use commonness and relatedness features
to rank a varying number of candidate senses of a particular mention
\cite{ref:Ferragina2010A, ref:Ferragina2010B, ref:Han2011B, ref:Milne2008, ref:Piccinno2014}.
Commonness features usually measure how common a sense is with respect 
to a mention, e.g. the probability that sense is used, given 
a particular mention. 
In our per-mention multi-class classification approach, all candidate-specific 
features that do not depend on the annotation context, including commonness, 
can be encapsulated in a single bias parameter for each class. 

Relatedness, measuring the pairwise similarity 
between candidate senses of multiple mentions in same documents,
is a very good coherence measurement of candidate senses. 
However, raletedness features heavily rely on the mention spotter, and have two possible shortcomings. Firstly, the spotter may fail
to recognize a good number of potential mentions due to the informal 
and short data, such as tweets or chats. The lack of 
extracted mentions leads to a weak relatedness measurement
and hence bad disambiguation. Secondly, a poorly trained spotter
might return low quality mentions such as ``yes'' and ``no'', while
even a good spotter may result in redundant or noisy mentions. 
Contrary to the belief that more mentions is better
for disambiguation, Piccinno showed that these noisy mentions
can reduce the disambiguation accuracy \cite{ref:Piccinno2014}
\footnote{The comparison between full D2W and light D2W \cite{ref:Piccinno2014}}.

So, instead of using relatedness features, we decide to 
use the light-weight and robust word-based similarity features
between annotation context and candidate sense context. 
We show that coupling the specialized per-mention classifier with
these features, which are tuned to contrast candidate senses,
can deliver a very accurate and fast disambiguation solution.

Usually, the similarity is measured by matching the
annotation context directly to contexts of candidate senses
such as the abstracts or the whole Wikipedia pages\cite{ref:Kulkarni2009}.
However, as the disambiguation is a per-mention 
classification, we can further tune candidate contexts 
to emphasize the contrast among candidates
of a mention before evaluating the word-based similarity.

Specifically, for all unique candidate senses $e^{(i)}$ with $(i=1, \ldots, |e|)$ 
across the entire Wikipedia dataset of 
a particular mention $m$, we construct the tf-idf matrix, select 
 and rank the top $n_{f;w}$ words by tf-idf values for each candidate, 
and put these ranked words into $n_{f;p}$ parts. 
As a result, for each candidate $e^{(i)}$, there are
$n_{f;p}$ context parts $(c^{(e;i)}_{1}, \ldots, c^{(e;i)}_{n_{f;p}})$;
each part is a list of word-and-tfidf pairs
$c^{(e;i)}_{j} = ((w^{(e;i)}_{j,1}, v^{(e;i)}_{j,1}), \ldots, (w^{(e;i)}_{j,\underline{n}_{f;w}}, v^{(e;i)}_{j,\underline{n}_{f;w}}) )$
where $w^{(e;\cdot)}_{\cdot,\cdot}$ is a ranked lemmatized word of an entity context, 
$v^{(e;\cdot)}_{\cdot,\cdot}$ is its corresponding tf-idf value and 
$\underline{n}_{f;w} = \lceil n_{f;w} / n_{f;p} \rceil$ is the number of words per part
\footnote{In the case that an entity $e^{(i)}$ has a short Wikipedia article body 
then some context parts $c^{(e;i)}_{j}$ may be empty or have less than $\underline{n}_{f;w}$ words}.
Due to the properties of idf components, the above procedure results in 
the contrasting contexts of different candidate senses.

For an unlabelled annotation $a$,
we transform its context $a_{context}$
to a list of words and counts 
$c^{(a)} = ((w^{(a)}_{1}, v^{(a)}_{1}), \ldots)$ where 
$w^{(a)}_{\cdot}$ is a unique word in the context $a_{context}$ and 
$v^{(a)}_{\cdot}$ is its occurrence count. The similarity  
between $a_{context}$ and a context part $c^{(e;i)}_{j}$ 
of a candidate sense $e^{(i)}$ are measured in four different ways:
\begin{align*}
  sim_{wo}(a,c^{(e;i)}_{j}) &= \frac{\sum_{k_1,k_2} \mathbbm{I}(w^{(a)}_{k_1}=w^{(e;i)}_{j,k_2})}{\log (|a_{context}|+1)}, \\
  sim_{ws}(a,c^{(e;i)}_{j}) &= \frac{\sum_{k_1,k_2} v^{(a)}_{k_1} \mathbbm{I}(w^{(a)}_{k_1}=w^{(e;i)}_{j,k_2})}{\log (|a_{context}|+1)}, \\
  sim_{to}(a,c^{(e;i)}_{j}) &= \frac{\sum_{k_1,k_2} v^{(e;i)}_{j,k_2} \mathbbm{I}(w^{(a)}_{k_1}=w^{(e;i)}_{j,k_2})}{\log (|a_{context}|+1)}, \\
  sim_{ts}(a,c^{(e;i)}_{j}) &= \frac{\sum_{k_1,k_2} v^{(a)}_{k_1} v^{(e;i)}_{j,k_2} \mathbbm{I}(w^{(a)}_{k_1}=w^{(e;i)}_{j,k_2})}{\log (|a_{context}|+1)},
\end{align*}
where $\mathbbm{I}(\cdot)$ is the indicator function and 
$|a_{context}|$ is the length of the annotation context, 
used for scaling different context length.
So, the feature vector to disambiguate an annotation $a$ 
of mention $m=a_{m}$ is $r^{(a)} = (r^{(a)}_1, \ldots, r^{(a)}_{n_{f;sim}})$,
which is an aggregation of all above similarities of all candidates senses $e^{(i)}$; 
$n_{f;sim} = 4 n_{f;p} |e|$ is the total number of features.

\subsection{Learning model}
\label{subsec:learning_model}

To perform per mention disambiguation, we use 
multinomial regression with the above similarity features, which 
is a statistically-proper classification model
\footnote{For this problem, we consider a classification model to be statistically
proper if it has a clear notion of statistical likelihood that
addresses multiple candidate senses per annotation observation. Note
that one-versus-the-rest logistic regression is not proper with
respect to this definition.}. 
Such a classification model
can provide a consistent predicted class probability, which 
is useful for later pruning.

For a labelled annotation $a$ with mention $m$, observed senses $e$
and features $r^{(a)}_{k}$, the multinomial regression model 
can be defined as follows:
\begin{align*}
  \mu^{(a)}_i &= \alpha_{i,0} + \sum_k \alpha_{i,k} r^{(a)}_k,\\
  p^{(a)}_i &= \frac{\exp(\mu^{(a)}_i)}{\sum_{i^{\prime}} \exp(\mu^{(a)}_{i^{\prime}})}, \\
  e^{(a)} &\sim p^{(a)}.
\end{align*}

For the disambiguation of new annotations after the  training phase, 
our method has the complexity $\mathcal{O}(|e|\times|a_{context}|)$ per-annotation.
The system is highly
scalable and all trained-model parameters, candidate senses, processing 
can be split by mentions evenly across different cluster nodes. 
The disambiguation process for any document can be parallelized 
on the annotation-level, which is important for real time processing of 
very long documents.


\section{Experiment Set-up}
\label{sec:ex_setup}

\subsection{System and implementation}

One of the numerical challenges for this approach is 
the required computation power needed for the processing
of more than 700K of unique ambiguous mentions and 170 million
labelled annotations. Fortunately, 
as the feature construction and classification learning is 
per-mention, the disambiguation system is highly compatible 
with a data-parallel computation system. 
So, in order to deal with the numerical computation, 
we use Apache Spark\footnote{http://spark.apache.org/}, a distributed processing system 
based on Map-Reduce 
framework, for all data processing, feature extraction and
model learning. As Spark supports 
distributed in-memory
serialization, caching and parallel processing for big data
with a simple but mature API, it satisfies our needs
and provides a nice speed up. 
Hence, a Spark cluster is set up for this disambiguation system,
consisting three 16$\times$2.6GHz 96GB-RAM machines. 

All the algorithms and procedures are implemented in Python
with PySpark API. For machine learning methods, we use 
the standard open source library scikit-learn.
Even though the code is not highly optimized like 
a C/C++ implementation, 
it is shown later that the system perform disambiguation 
in about 2.8ms per annotation.

\subsection{Training and validation set-up}
\label{subsec:training_validation_set_up}

For the purposes of training and validation, 
the annotation dataset $\mathcal{A}$ in Section \ref{sec:data}
is split by ratio (90\%,10\%) per-mention. The 90\% training
dataset is denoted by $\mathcal{A}_1$ and the other is by $\mathcal{A}_2$.
In order to validate the disambiguation system 
in different data scenarios such as short-text 
and noisy-text, we use the following
transformation on the original annotation dataset 
and create different datasets.

For a mention $m$ and
candidate senses $e^{(i)}$ $(i=1, \ldots, |e|)$, 
a random noisy vocabulary $\mathcal{V}_{m}$
of unique words is first constructed 
by doing set-union on words of $e^{(i)}_{body}$.
Then for every annotation $a$ of mentions $m$, a new 
annotation context $a^{\prime}_{context}$ is formed 
by sampling-without-replacement 
$n_{s,1} = p_{s,1} \times |a_{context}|$ words from $a_{context}$
and $n_{s,2} = p_{s,2} \times |a_{context}|$ words from $\mathcal{V}_{m}$.
All new annotations $a^{\prime}$ with contexts $a^{\prime}_{context}$
form a new dataset. Parameter $p_{s,1} \leq 1$ 
is the shrinkage factor of the transformation while 
parameter $p_{s,2}$ 
implies the noise level from other candidate senses.

Four such datasets are constructed with parameters $p_{s,1}$, 
$p_{s,2}$ specified in Table \ref{tab_trvld:data_transform_par}
and are only used for validation purpose. Notice that 
any n-gram mentions ($n\geq 2$) of the original annotation
context can be broken by the sampling operation. 
Our disambiguation system, by design, is robust with
respect to such scrambled contexts while relatedness-based
methods would suffer from this added noise.

\begin{table}[t]
\caption{Data transformation parameters}
\centering
\begin{tabular}{|c|c|c|c|c|}
\hline
Dataset &$\mathcal{B}$ &$\mathcal{C}$ &$\mathcal{D}$ &$\mathcal{E}$\\
\hline
$p_{s,1}$ &80\% &60\% &40\% &20\%\\
\hline
$p_{s,2}$ &20\% &0\% &0\% &0\%\\
\hline
\end{tabular}
\label{tab_trvld:data_transform_par}
\end{table}

\subsection{Metrics}

For comparison purpose, we use the following
precision and recall definitions: 
\begin{align*}
  \mathcal{P} &= \frac{ \sum_{d_i,a \in d_i, \widetilde{a} \in d_i} \mathbbm{I}(a=\widetilde{a}, a_e=\widetilde{a}_e) }{ \sum_{d_i, \widetilde{a} \in d_i} 1 },\\
  \mathcal{R} &= \frac{ \sum_{d_i,a \in d_i, \widetilde{a} \in d_i} \mathbbm{I}(a=\widetilde{a}, a_e=\widetilde{a}_e) }{ \sum_{d_i, a \in d_i} 1 },
\end{align*}
where $d_i$ is a document; $a$ and $\widetilde{a}$ are ground-truth annotation 
and predicted annotations, correspondingly
\footnote{Two annotations $a$ and $\widetilde{a}$ are equal 
when they are in the same location and in the same source document}.
These definitions are equivalent to the comparison metrics
in the long term track in ERD-2014's challenge
\cite{ref:Carmel2014}.

The definitions of precision and recall above may 
be biased to mentions with a large number of labelled annotations
in Wikipedia dataset. Hence, we also use 
the following average precision-recall across mentions:
\begin{align*}
  \underline{\mathcal{P}} &= \frac{\sum_m P_m}{|m|}, \\
  \underline{\mathcal{R}} &= \frac{\sum_m R_m}{|m|},
\end{align*}
where $|m|$ is the number of unique mentions; 
$P_m$ and $R_m$ are precision and recall of a specific mention $m$.

\section{Analysis on Learning Settings}
\label{sec:analysis01}

In this section, we explore and analyze the accuracy 
of the proposed disambiguation system by varying 
several configurable variables. 

In the feature extraction step,
$n_{f;w}$ defines the number of unique words, ranked by 
tfidf values, in each candidate sense context, 
used for matching with an annotation context.
In the case of using a large value of $n_{f;w}$, 
we may expect the effect of high ranking words to
the disambiguation classifier is different from ones
of low ranking words, and hence divide them in a number of
parts $n_{f;p}$, as described in Section \ref{subsec:feature_extraction}. 
In terms of computation, $n_{f;w}$ affects 
the cost of matching the annotation context 
with the top-ranked words of candidate context 
while $n_{f;p}$ affects the number of training features.

Another variable that affects the system performance
is the classifier. So, aside from multinomial regression, 
we also evaluate the accuracy results with 
one-versus-the-rest logistic regression, 
random forest and support vector classification.

For this analysis of configurable system variables, 
the system is trained and evaluated on 8834 random unique 
mentions; also in any dataset, 
if there are more than $5000$ annotations 
of a specific candidate sense of a mention, we sample-without-replacement
$5000$ annotations for that pair of mention-sense 
to reduce the experiment running time
\footnote{As running the system on the entire Wikipedia data
would take several days for just one setting, even with the
help of the Spark cluster, we run this
setting analysis on a smaller random subset first 
and provide a full-run performance for one setting in Section
\ref{sec:full_exp}}. The validation results are provided
for both the original validation dataset $\mathcal{A}_2$
and the scrambled datasets in 
Section \ref{subsec:training_validation_set_up}.

\begin{table*}[t]
\caption{Performance results of different settings $(n_{f;w},n_{f;p})$ with multinomial regression. The best results are in bold.}
\centering
\begin{adjustbox}{center}
\begin{tabular}{|c|c|c|c|c|c|c|c|c|c|c|c|c|c|}
\hline
\multirow{ 2}{*}{$n_{f;w}$} 
&\multirow{ 2}{*}{$n_{f;p}$} 
&\multirow{ 2}{*}{$\mathcal{P}^{\mathcal{A}_2}$}
&\multirow{ 2}{*}{$\underline{\mathcal{P}}^{\mathcal{A}_2}$}
&\multirow{ 2}{*}{$\mathcal{P}^{\mathcal{B}}$}
&\multirow{ 2}{*}{$\underline{\mathcal{P}}^{\mathcal{B}}$}
&\multirow{ 2}{*}{$\mathcal{P}^{\mathcal{C}}$}
&\multirow{ 2}{*}{$\underline{\mathcal{P}}^{\mathcal{C}}$}
&\multirow{ 2}{*}{$\mathcal{P}^{\mathcal{D}}$}
&\multirow{ 2}{*}{$\underline{\mathcal{P}}^{\mathcal{D}}$}
&\multirow{ 2}{*}{$\mathcal{P}^{\mathcal{E}}$}
&\multirow{ 2}{*}{$\underline{\mathcal{P}}^{\mathcal{E}}$}
&$\mathcal{T}_{\text{total}}$ &$\mathcal{T}_{\text{pred}}$\\

&
&&
&&
&&
&&
&&
&$(\times10^3 \text{s})$ &$(\text{ms})$\\
\hline
400 &8 &\textbf{.9186} &.9206 &\textbf{.9325} &\textbf{.9274} &\textbf{.9351} &\textbf{.9529} &\textbf{.9053} &\textbf{.9260} &\textbf{.8550} &\textbf{.8787} &47.56 &5.69\\
\hline
300 &6 &.9182 &.9199 &.9315 &.9266 &.9331 &.9504 &.9031 &.9231 &.8536 &.8771 &39.18 &4.89\\
\hline
200 &4 &.9174 &.9200 &.9297 &.9248 &.9298 &.9458 &.9005 &.9183 &.8524 &.8744 &29.66 &3.95\\
\hline
100 &2 &.9157 &.9163 &.9243 &.9203 &.9225 &.9347 &.8947 &.9098 &.8487 &.8686 &2.55 &3.00\\
\hline
400 &1 &.9152 &\textbf{.9215} &.9213 &.9186 &.9182 &.9296 &.8951 &.9106 &.8532 &.8754 &24.32 &3.91\\
\hline
300 &1 &.9150 &.9211 &.9212 &.9181 &.9181 &.9292 &.8949 &.9104 &.8530 &.8751 &22.74 &3.61\\
\hline
200 &1 &.9147 &.9203 &.9208 &.9175 &.9174 &.9286 &.8940 &.9095 &.8518 &.8740 &2.71 &3.33\\
\hline
100 &1 &.9138 &.9188 &.9193 &.9163 &.9160 &.9263 &.8916 &.9063 &.8491 &.8701 &\textbf{18.13} &\textbf{2.81}\\
\hline
\end{tabular}
\end{adjustbox}
\label{tab_analysis01:nword_npart}

\end{table*}

First, performance results by varying $n_{f;w}$ and $n_{f;p}$
with multinomial regression are given in 
Table \ref{tab_analysis01:nword_npart}.
$\mathcal{T}_{\text{total}}$ is the total time of feature 
construction, training and validation  of all datasets and 
$\mathcal{T}_{\text{pred}}$ is the prediction time per-annotation 
(including the feature construction time); 
both are measured in a sequential manner as the running time 
of all mentions in all Spark executor instances is summed up 
before the evaluation.

As we want to validate purely the disambiguation process, 
we do not prune off uncertain predictions in this section
and the disambiguation always returns a non-NIL candidate for
any annotation. Consequently, precision, recall and F-measure 
are all equal and only precision values are reported. 

We make the following observations about Table \ref{tab_analysis01:nword_npart}:
\begin{itemize}
\item Increasing $n_{f;w}$ and $n_{f;p}$ raises the precision but the 
increment magnitude is diminishing.
\item There is a trade off between precision and running time/prediction time. The more the number of top-ranked candidate context words and 
the number of features, the higher the precision but 
the slower the disambiguation process and 
the longer the prediction time per-annotation.
\item As expected, the precision decreases 
when the annotation context becomes smaller and smaller from 
validation dataset $\mathcal{C}$ to $\mathcal{E}$. 
\item Between dataset $\mathcal{B}$ and $\mathcal{C}$, $\mathcal{B}$ has longer but noisier context than $\mathcal{C}$, resulting in a lower precision.
\item Mention-average precision $\underline{\mathcal{P}}$ 
is higher than the standard precision $\mathcal{P}$, in general,
implying that mentions with higher number of labelled annotations
have lower per-mention precision. Notice that these mentions
also have a much higher number of candidate senses, 
as shown in Figure \ref{fig_data:sense_annotation}. 
\end{itemize}

The comparison between different classifiers is shown in 
Table \ref{tab_analysis01:cls}. For this problem, 
in terms of precision, multinomial regression (MR)
and one-versus-the-rest logistic regression (LR) seem to be 
clear winners. However, LR and also support vector classifier (SVC) 
have very long training time and hence total time,
which is a big disadvantage when the disambiguation training
is applied on the full Wikipedia dataset. 
Accuracy results of random forest with 10 estimators (RF-10) or 30 estimators (RF-30)
are not as high as the others.
Hence, MR seems to be a better choice for this task,
especially when it is statistically proper and can provide
the coherent prediction probability across candidate classes
for the pruner described in the next section.

\begin{table*}[t]
\caption{Performance results of different classifiers (CLS) on two settings of $(n_{f;w},n_{f;p})$. The best results in each setting are in bold.}
\centering
\begin{adjustbox}{center}
\begin{tabular}{|c|c|c|c|c|c|c|c|c|c|c|c|c|c|}
\hline
$(n_{f;w},$
&\multirow{ 2}{*}{CLS} 
&\multirow{ 2}{*}{$\mathcal{P}^{\mathcal{A}_2}$}
&\multirow{ 2}{*}{$\underline{\mathcal{P}}^{\mathcal{A}_2}$}
&\multirow{ 2}{*}{$\mathcal{P}^{\mathcal{B}}$}
&\multirow{ 2}{*}{$\underline{\mathcal{P}}^{\mathcal{B}}$}
&\multirow{ 2}{*}{$\mathcal{P}^{\mathcal{C}}$}
&\multirow{ 2}{*}{$\underline{\mathcal{P}}^{\mathcal{C}}$}
&\multirow{ 2}{*}{$\mathcal{P}^{\mathcal{D}}$}
&\multirow{ 2}{*}{$\underline{\mathcal{P}}^{\mathcal{D}}$}
&\multirow{ 2}{*}{$\mathcal{P}^{\mathcal{E}}$}
&\multirow{ 2}{*}{$\underline{\mathcal{P}}^{\mathcal{E}}$}
&$\mathcal{T}_{\text{total}}$ &$\mathcal{T}_{\text{pred}}$\\

~$n_{f;p})$
&
&&
&&
&&
&&
&&
&$(\times10^3 \text{s})$ &$(\text{ms})$\\

\hline
\multirow{ 4}{*}{(400,8)} &SVC &.9019 &.8185 &.9237 &.9029 &.9269 &.9375 &.8935 &.9037 &.8397 &.8495 &228.46 &5.85\\
\cline{2-14}
 &MR &\textbf{.9186} &.9206 &.9325 &\textbf{.9274} &.9351 &\textbf{.9529} &.9053 &\textbf{.9260} &.8550 &\textbf{.8787} &47.56 &5.69\\
\cline{2-14}
 &LR &.9127 &.8147 &\textbf{.9350} &.9091 &\textbf{.9362} &.9389 &\textbf{.9091} &.9115 &\textbf{.8610} &.8644 &252.24 &5.70\\
\cline{2-14}
 &RF-10 &.8938 &.9165 &.8889 &.8838 &.8828 &.9060 &.8563 &.8784 &.8223 &.8406 &\textbf{33.45} &\textbf{5.57}\\
\cline{2-14}
 &RF-30 &.9052 &\textbf{.9293} &.9039 &.9004 &.8972 &.9217 &.8709 &.8939 &.8368 &.8535 &35.15 &5.70\\
\hline
\hline
\multirow{ 4}{*}{(100,1)} &SVC &.9091 &.8979 &.9166 &\textbf{.9164} &.9140 &\textbf{.9286} &.8878 &.9030 &.8422 &.8595 &31.35 &2.86\\
\cline{2-14}
 &MR &.9138 &.9188 &\textbf{.9193} &.9163 &\textbf{.9160} &.9263 &.8916 &\textbf{.9063} &.8491 &\textbf{.8701} &18.13 &2.81\\
\cline{2-14}
 &LR &\textbf{.9140} &.9002 &.9185 &.9111 &.9154 &.9207 &\textbf{.8947} &.9015 &\textbf{.8560} &.8658 &43.41 &\textbf{2.78}\\
\cline{2-14}
 &RF-10 &.8989 &.9118 &.8966 &.8921 &.8884 &.9042 &.8632 &.8802 &.8288 &.8431 &\textbf{15.71} &2.80\\
\cline{2-14}
 &RF-30 &.9080 &\textbf{.9208} &.9089 &.9036 &.9001 &.9164 &.8746 &.8905 &.8402 &.8542 &16.65 &2.85\\
\hline
\end{tabular}
\end{adjustbox}
\label{tab_analysis01:cls}
\end{table*}


\section{Pruner}
\label{sec:pruning_pr}

In the previous section, 
for a pure analysis of disambiguation accuracy, 
the system is forced to return one candidate sense
for any annotation. In this section, we propose
two approaches to prune off uncertain annotation 
and explore the trade off between
precision and recall. These pruners are hence 
natural solutions for NIL-detection problem.

\subsection{Pruning with binary classifiers}

Similar to other proposed pruning solutions 
\cite{ref:Ferragina2010A,ref:Olieman2014,ref:Piccinno2014}
we also consider a binary classifier to prune off 
uncertain results from the previous disambiguation phase
in Section \ref{sec:disambiguation}.
However, instead of using a single global classifier, 
we faithfully follow a per-mention learning approach
and use a different binary classifier for each unique mention.

For any target ground-truth annotation $a$ of a unique mention $m$, 
a predicted annotation $\widetilde{a}$ is 
returned by the disambiguation. 
To learn the pruner of mention $m$,
three features are first evaluated for each $\widetilde{a}$: 
$g^{(\widetilde{a})}_{1}$ is the predicted probability
of multinomial classifier of $\widetilde{a}$; 
$g^{(\widetilde{a})}_{2}$ is the longest-common-string length
between $m$ and the predicted sense title $\widetilde{e}_{title}$ 
($\widetilde{e}=\widetilde{a}_e$); 
$g^{(\widetilde{a})}_{3}$ is another string-overlapping metric,
defined by the ratio 
$\frac{m \cap \widetilde{e}_{title}}{m \cup \widetilde{e}_{title}}$.
Both the length $g^{(\widetilde{a})}_{2}$ and the union, intersection
operators are measured on the word-level, not character-level;
disambiguation parts in Wikipedia title, e.g. 
the part "(language)" in "Java\_(language)", are removed before
the intersection and union evaluation. 
The objective of this pruner is to learn the binary label 
$\mathbbm{I}(a_e=\widetilde{a}_e)$ of a particular mention $m$.
Any new predicted annotation $\widetilde{a}^{\prime}$ is removed 
if its predicted binary pruning label is negative.

In order to avoid validation bias, we first employ the same 
technique in Section \ref{subsec:training_validation_set_up}
to create a new dataset $\mathcal{F}$ with a non-zero $p^{(\mathcal{F})}_{s,1}$ 
(noise is not considered for this dataset: $p^{(\mathcal{F})}_{s,2}=0$)
and then train a binary classifier with $\mathcal{F}$. 
Finally, the trained classifiers are applied on the disambiguation 
results of other validation datasets 
such as $\mathcal{A}_2$, $\mathcal{B}$, etc. 

We further extend the pruning analysis by using 
a binary classifier on each candidate of a mention. 
In this case, features 
$g^{(\widetilde{a})}_{2:3}$ are dropped as they are 
mathematically equivalent to a single per-candidate 
bias parameter. 

The pruning results for both approaches are shown in
Table \ref{tab_analysis02:pr_binary_classifier}:
the classifier used for these experiments is 
random forest with 30 estimators; 
$p^{(\mathcal{F})}_{s,1}$ is set to 80\%.
$\mathcal{P}$ is the precision (the same as recall and F-measure)
before pruning while
$\mathcal{P}_{\text{pr}}$, $\mathcal{R}_{\text{pr}}$, 
$\mathcal{F}_{\text{pr}}$ are
the precision, recall, F-measure after pruning. 
From the table, 
it can be seen that the precision can be pushed up 
to 0.5\%-1.0\% higher, 
with the penalty of recall (decreased by 2.8\%-3.0\%); 
also, the difference between per-mention 
and per-candidate classifiers is negligible.
The pruning effect by using a binary classifier
on setting $(n_{f;w}=\text{100}, \allowbreak n_{f;p}=\text{1})$
is a little bit stronger than on 
$(n_{f;w}=\text{400}, \allowbreak n_{f;p}=\text{8})$: 
higher precision (increased by 0.7\%-1.3\%) 
but lower recall (decreased by 4.3\%-4.7\%)

We also conduct other pruning experiments 
using logistic regression instead of random forest.
However, it has a very little pruning effect 
(all metrics only vary about 0.2\%-0.3\%),
implying a similar classification effect
between the disambiguation classifier MR 
and the pruner classifier LR.

\begin{table*}[t]
\caption{Precision-recall trade-off results with binary classifiers}
\begin{adjustbox}{center}
\begin{tabular}{|c|c|c|c|c|c|c|c|c|}
\hline
\multirow{2}{*}{$n_{f;w}$}
&\multirow{2}{*}{$n_{f;p}$}
&Pruner
&\multirow{2}{*}{Metric}
&\multicolumn{5}{c|}{Dataset}\\
\cline{5-9}
&
&type
&
&$\mathcal{A}_2$
&$\mathcal{B}$
&$\mathcal{C}$
&$\mathcal{D}$
&$\mathcal{E}$\\
\hline
\multirow{7}{*}{400} &\multirow{7}{*}{8} & &$\mathcal{P}$ &.9186 &.9325 &.9351 &.9053 &.8550\\
\cline{3-9}
 & &\multirow{ 3}{*}{per-mention} &$\mathcal{P}_{\text{pr}}$ &.9267 &.9379 &.9404 &.9119 &.8623\\
\cline{4-9}
 & & &$\mathcal{R}_{\text{pr}}$ &.8897 &.9028 &.9054 &.8760 &.8265\\
\cline{4-9}
 & & &$\mathcal{F}_{\text{pr}}$ &.9079 &.9200 &.9226 &.8936 &.8440\\
\cline{3-9}
 & &\multirow{ 3}{*}{per-candidate} &$\mathcal{P}_{\text{pr}}$ &.9288 &.9386 &.9412 &.9127 &.8634\\
\cline{4-9}
 & & &$\mathcal{R}_{\text{pr}}$ &.8901 &.9024 &.9054 &.8759 &.8267\\
\cline{4-9}
 & & &$\mathcal{F}_{\text{pr}}$ &.9091 &.9202 &.9229 &.8939 &.8446\\
\hline
\multirow{7}{*}{100} &\multirow{7}{*}{1} & &$\mathcal{P}$ &.9138 &.9193 &.9160 &.8916 &.8491\\
\cline{3-9}
 & &\multirow{ 3}{*}{per-mention} &$\mathcal{P}_{\text{pr}}$ &.9238 &.9269 &.9240 &.9000 &.8573\\
\cline{4-9}
 & & &$\mathcal{R}_{\text{pr}}$ &.8702 &.8727 &.8700 &.8465 &.8064\\
\cline{4-9}
 & & &$\mathcal{F}_{\text{pr}}$ &.8962 &.8990 &.8962 &.8724 &.8311\\
\cline{3-9}
 & &\multirow{ 3}{*}{per-candidate} &$\mathcal{P}_{\text{pr}}$ &.9266 &.9276 &.9248 &.9007 &.8580\\
\cline{4-9}
 & & &$\mathcal{R}_{\text{pr}}$ &.8696 &.8727 &.8703 &.8472 &.8066\\
\cline{4-9}
 & & &$\mathcal{F}_{\text{pr}}$ &.8972 &.8993 &.8967 &.8731 &.8315\\
\hline
\end{tabular}
\end{adjustbox}
\label{tab_analysis02:pr_binary_classifier}

\end{table*}

\subsection{Thresholding predicted probability}

A binary classifier in the previous section is a nice and
automatic pruner. However, in some cases, we still want to 
increase the pruning strength, 
pushing precision even higher at the cost of recall;
hence, in this section, we use a method to adjust the 
lower threshold of the
predicted probability: only predicted annotation with 
$g^{(\widetilde{a})}_{1}$ higher than the threshold is
kept by the system. Instead of using a global 
lower threshold value, we find one threshold value
for each candidate of a mention, satisfying
a global predefined condition of F-measure and precision. 

The adjustment is trained with dataset $\mathcal{F}$
as in the previous section. For each mention, all 
predicted annotations are grouped accordingly to 
their predicted senses $\widetilde{e}$ and then 
Algorithm \ref{alg:pruner_adj} is applied on each group independently.
Finally, the estimated thresholds are used to prune
off uncertain annotations in the other datasets
such as $\mathcal{A}_2$, $\mathcal{B}$, etc.

The pruning results of Algorithm \ref{alg:pruner_adj}
are shown in Table \ref{tab_analysis02:pr_pc_adj}.
From both the algorithm and the results, it 
can be seen that 
the lower the control parameters $\beta_0$, $\beta_1$,
the looser the condition and the higher the potential precision
value. The procedure consistently increases
the precision values by 1.1\%-2.8\% 
across different settings and datasets, 
illustrating a stronger pruning effect than using binary classifiers.

In conclusion for this section, 
we believe pruning is very important as it increases the precision 
and provides non-noisy disambiguation results needed for subsequent 
text analysis such as summarization or similarity measurement.
It is especially true in our system as the 
recall value is quite high  
and we can extract enough information from the ground truth.

\begin{algorithm}[!ht]
\caption{Per-candidate threshold adjustment}
\label{alg:pruner_adj}

\KwIn{Predicted annotations $\widetilde{a}^{(j)}$ ($j=1 \ldots J$) of the same
predicted sense $\widetilde{e}$}
\nonl \myinput{Each annotation $\widetilde{a}^{(j)}$ is associated with 
a predicted probability $g^{(j)}=g^{(\widetilde{a}^{(j)})}_{1}$
and a binary label $h^{(j)}$ ($h^{(j)}=1$ if
$\widetilde{a}^{(j)}$ is a correct annotation)}
\nonl \myinput{Control parameters $\beta_0$ and $\beta_1$}

Calculate the precision $p^{\prime}$ and F-measure $f^{\prime}$ when there is no pruning threshold,
using labels $h^{(j)}$.

Calculate the precision $p^{(j)}$, F-measure $f^{(j)}$ at all threshold values $g^{(j)}$.

Find the index set 
$\mathcal{S}=\{k: (f^{(k)} - f^{\prime}) \geq \beta_0, 
\text{and } (p^{(k)} - p^{\prime} + f^{(k)} - f^{\prime}) \geq \beta_1 \}$. 

If $\mathcal{S}$ not empty, find $k^{\prime} = \argmax_{k \in \mathcal{S}}(p^{(k)})$.
Otherwise, find $k^{\prime} = \argmax_{j=1 \ldots J}(f^{(j)}, p^{(j)})$.

Return $g^{(k^{\prime})}$ as the target threshold.
\end{algorithm}

\begin{table*}[t]
\caption{Precision-recall trade-off results with per-candidate threshold adjustment}
\begin{adjustbox}{center}
\begin{tabular}{|c|c|c|c|c|c|c|c|c|c|}
\hline
\multirow{2}{*}{$n_{f;w}$}
&\multirow{2}{*}{$n_{f;p}$}
&\multirow{2}{*}{$\beta_0$}
&\multirow{2}{*}{$\beta_1$}
&\multirow{2}{*}{Metric}
&\multicolumn{5}{c|}{Dataset}\\
\cline{6-10}
&
&
&
&
&$\mathcal{A}_2$
&$\mathcal{B}$
&$\mathcal{C}$
&$\mathcal{D}$
&$\mathcal{E}$\\

\hline
\multirow{8}{*}{400} &\multirow{8}{*}{8} & & &$\mathcal{P}$ &.9186 &.9325 &.9351 &.9053 &.8550\\
\cline{3-10}
 & &\multirow{3}{*}{-0.05} &\multirow{3}{*}{-0.02} &$\mathcal{P}_{\text{pr}}$ &.9327 &.9439 &.9472 &.9206 &.8735\\
\cline{5-10}
 & & & &$\mathcal{R}_{\text{pr}}$ &.8911 &.9022 &.9058 &.8735 &.8205\\
\cline{5-10}
 & & & &$\mathcal{F}_{\text{pr}}$ &.9114 &.9226 &.9261 &.8964 &.8462\\
\cline{3-10}
 & &\multirow{3}{*}{-0.15} &\multirow{3}{*}{-0.05} &$\mathcal{P}_{\text{pr}}$ &.9374 &.9479 &.9512 &.9257 &.8802\\
\cline{5-10}
 & & & &$\mathcal{R}_{\text{pr}}$ &.8691 &.8780 &.8820 &.8498 &.7974\\
\cline{5-10}
 & & & &$\mathcal{F}_{\text{pr}}$ &.9019 &.9116 &.9153 &.8861 &.8367\\
\hline
\multirow{8}{*}{100} &\multirow{8}{*}{1} & & &$\mathcal{P}$ &.9138 &.9193 &.9160 &.8916 &.8491\\
\cline{3-10}
 & &\multirow{3}{*}{-0.05} &\multirow{3}{*}{-0.02} &$\mathcal{P}_{\text{pr}}$ &.9301 &.9343 &.9321 &.9095 &.8690\\
\cline{5-10}
 & & & &$\mathcal{R}_{\text{pr}}$ &.8769 &.8777 &.8763 &.8499 &.8058\\
\cline{5-10}
 & & & &$\mathcal{F}_{\text{pr}}$ &.9027 &.9051 &.9033 &.8787 &.8362\\
\cline{3-10}
 & &\multirow{3}{*}{-0.15} &\multirow{3}{*}{-0.05} &$\mathcal{P}_{\text{pr}}$ &.9358 &.9399 &.9377 &.9162 &.8766\\
\cline{5-10}
 & & & &$\mathcal{R}_{\text{pr}}$ &.8448 &.8427 &.8423 &.8160 &.7734\\
\cline{5-10}
 & & & &$\mathcal{F}_{\text{pr}}$ &.8879 &.8886 &.8875 &.8632 &.8218\\
\hline
\end{tabular}
\end{adjustbox}
\label{tab_analysis02:pr_pc_adj}

\end{table*}


\section{Comparison to Other Systems}
\label{sec:spotlight}

\begin{table}[t]
\caption{Comparison of Dbpedia Spotlight (DS) and our proposed system (PML)}
\begin{adjustbox}{center}
\begin{tabular}{|c|c|c|c|c|c|}
\hline
DS instance
&\multirow{2}{*}{$|{\mathcal{G}^{\prime}}|$}
&\multirow{2}{*}{$\mathcal{P}_{DS}$}
&\multirow{2}{*}{$\underline{\mathcal{P}}_{DS}$}
&\multirow{2}{*}{$\mathcal{P}_{PML}$}
&\multirow{2}{*}{$\underline{\mathcal{P}}_{PML}$}\\
($\gamma$)
&
&
&
&
&\\

\hline
0.0 &65109 &.8781 &.8169 &.9035 &.8985\\
\hline
0.5 &64586 &.8822 &.8201 &.9051 &.8989\\
\hline
\end{tabular}
\end{adjustbox}
\label{tab_analysis03:comp_dbpedia_spotlight}
\end{table}
\begin{table}[t]
\caption{Comparison of TagMe (TM) and our proposed system (PML)}
\begin{adjustbox}{center}
\begin{tabular}{|c|c|c|c|c|}
\hline
$|{\mathcal{G}^{\prime}}|$ & $\mathcal{P}_{TM}$ & $\underline{\mathcal{P}}_{TM}$ & $\mathcal{P}_{PML}$ & $\underline{\mathcal{P}}_{PML}$ \\
\hline
37872 &.8752 &.8244 &.9077 &.8950 \\
\hline
\end{tabular}
\end{adjustbox}
\label{tab_analysis03:comp_tagme}
\end{table}

\newcolumntype{R}[2]{
    >{\adjustbox{angle=#1,lap=\width-(#2)}\bgroup}
    l
    <{\egroup}
}
\newcommand*\rotb[1]{\multicolumn{1}{c}{\rotatebox{90}{#1}}}

\newcommand*\rot{\multicolumn{1}{R{65}{1em}}}

\newcommand*\textRankOne[1]{{\color{red}\textbf{$\dagger$#1}}}
\newcommand*\textRankTwo[1]{{\color{blue}\textbf{$\ddagger$#1}}}

\begin{table}[p]
\caption{\label{tab:gerbil} GERBIL comparison of different systems.  The micro-F1 (top) and macro-F1 (bottom) scores of each system on each dataset are reported.  Each column displays the best micro/macro-F1 score in red (marking the row with $\dagger$), and the second best micro/macro-F1 score in blue (marking the row with $\ddagger$).}
\centering
\begin{tabular}{l||r|r|r|r|r|r|r|r|r|r|r}
 & \multicolumn{11}{c}{Datasets} \\
                     & \rotb{ACE2004} & \rotb{AIDA-CoNLL} & \rotb{AQUAINT} & \rotb{DBSpotlight} & \rotb{IITB} & \rotb{KORE50} & \rotb{Micropost} & \rotb{MSNBC} & \rotb{N3-Reuters-128} & \rotb{N3-RSS-500} & \rotb{OKE-2015} \\
\hline
\hline
\multirow{2}{*}{PML}       & .637 & \textRankTwo{.545} & .685 & \textRankOne{.806} & .460 & .403 & .527 & .573 & \textRankTwo{.553} & \textRankOne{.677} & .737 \\ 
                           &\textRankOne{.793} & \textRankTwo{.571} & .683 & \textRankOne{.812} & \textRankTwo{.459} & .376 & .729 & \textRankOne{.648} & \textRankTwo{.592} & \textRankOne{.676} & .742 \\ 
\hline
\hline
\multirow{2}{*}{AGDISTIS}  & .618 & .498 & .508 & .263 & \textRankTwo{.467} & .323 & .323 & \textRankTwo{.621} & \textRankOne{.642} & \textRankTwo{.607} & .615 \\ 
                           & .752 & .491 & .495 & .273 & \textRankOne{.480} & .290 & .593 & .569 & \textRankOne{.699} & \textRankTwo{.607} & .629 \\ 
\hline
\multirow{2}{*}{AIDA}      & .076 & .416 & .071 & .210 & .166 & \textRankTwo{.623} & .331 & .069 & .353 & .404 & .617 \\ 
                           & .410 & .384 & .072 & .184 & .173 & \textRankTwo{.563} & .556 & .077 & .294 & .347 & .607 \\ 
\hline
\multirow{2}{*}{Babelfy}   & .517 & .543 & .668 & .520 & .364 & \textRankOne{.731} & .471 & .600 & .439 & .441 & .684 \\ 
                           & .685 & .496 & .667 & .512 & .348 & \textRankOne{.696} & .621 & .538 & .378 & .379 & .663 \\ 
\hline
\multirow{2}{*}{DBSpotlight} & .471 & .426 & .520 & .701 & .296 & .439 & .495 & .351 & .325 & .200 & .244 \\ 
                           & .664 & .436 & .502 & .675 & .279 & .401 & .660 & .333 & .255 & .161 & .200 \\ 
\hline
\multirow{2}{*}{Dexter}    & .507 & .407 & .513 & .284 & .204 & .183 & .404 & .293 & .354 & .369 & .580 \\ 
                           & .667 & .387 & .502 & .251 & .204 & .123 & .587 & .298 & .302 & .293 & .510 \\ 
\hline
\multirow{2}{*}{EC-NER}    & .488 & .439 & .403 & .244 & .137 & .290 & .412 & .429 & .365 & .331 & .192 \\ 
                           & .656 & .420 & .369 & .194 & .150 & .252 & .594 & .407 & .335 & .320 & .160 \\ 
\hline
\multirow{2}{*}{Kea}       & .634 & .539 & \textRankOne{.763} & \textRankTwo{.733} & \textRankOne{.472} & .588 &\textRankOne{.631} & \textRankOne{.662} & .501 & .435 &\textRankTwo{.761} \\ 
                           & .755 & .524 & \textRankOne{.753} & \textRankTwo{.725} & .453 & .527 & \textRankOne{.758} & \textRankTwo{.615} & .447 & .387 & \textRankTwo{.753} \\ 
\hline
\multirow{2}{*}{NERD-ML}   & .558 & .465 & .575 & .548 & .422 & .312 & .478 & .513 & .402 & .367 & .740 \\ 
                           & .714 & .427 & .554 & .528 & .411 & .252 & .629 & .502 & .340 & .297 & .719 \\ 
\hline
\multirow{2}{*}{TagMe 2}   & \textRankOne{.660} & .513 & \textRankTwo{.723} & .661 & .385 & .590 & .578 & .590 & .445 & .470 & \textRankOne{.832} \\ 
                           & \textRankTwo{.776} & .481 & .708 & .642 & .372 & .532 & .712 & .556 & .380 & .391 & \textRankOne{.814} \\ 
\hline
\multirow{2}{*}{WAT}       & \textRankTwo{.643} & \textRankOne{.597} & .714 & .653 & .401 & .593 & \textRankTwo{.601} & .601 & .504 & .433 & .697 \\ 
                           & .758 & \textRankOne{.581} & \textRankTwo{.714} & .666 & .385 & .491 & \textRankTwo{.740} & .542 & .427 & .364 & .648 \\ 
\end{tabular}
\end{table}

For the remainder of the paper we refer to the \emph{per-mention learning} disambiguation system described above as \emph{PML}, for brevity.

\subsection{Comparison with Wikipedia as Ground Truth}

In this section, we compare the proposed disambiguation system with Dbpedia Spotlight\footnote{We used Spotlight 0.7~\cite{ref:Daiber2013} (statistical model \texttt{en\_2+2} with the SpotXmlParser.} and
TagMe\footnote{We used the TagMe version 1.8 web API \url{tagme.di.unipi.it/tag} in January, 2016.}.  In the experiment, we only used ground truth annotations from dataset $\mathcal{A}_2$; for fairness we do not use any annotations created by extending the data. 

An annotation set $\mathcal{G} \subset \mathcal{A}_2$ is used as an input of two Spotlight instances of different confidence values $\gamma=0.0$ and $\gamma=0.5$. 
We note that as Spotlight may not return disambiguation results for intended target mentions in annotations input due to pruning, Spotlight outputs are only for a subset $\mathcal{G}^{\prime} \subset \mathcal{G}$. 
We then use the proposed PML disambiguation system of setting $(n_{f;w}=100,n_{f;p}=1)$ without pruning on $\mathcal{G}^{\prime}$.
The precision results are shown in Table \ref{tab_analysis03:comp_dbpedia_spotlight}, indicating that our proposed system has a higher accuracy of between 2.2\% and 8.2\% depending on the metric. 
The precision drop from $\mathcal{P}_{DS}$ to $\underline{\mathcal{P}}_{DS}$ implies that Spotlight disambiguation does not work as well as PML across distinct mentions.

For TagMe, a similar methodology is employed, but with a minor difference: the TagMe web API does not allow the user to specify the annotation for disambiguation. 
As a result, we rely on the TagMe spotter, and only include results where TagMe annotated exactly the same mention as the ground truth data.
The precision results are shown in Table \ref{tab_analysis03:comp_tagme}, indicating that our proposed system has a higher accuracy from 3.3\% to~7.1\%.

\subsection{Comparison using GERBIL}
To provide convincing evidence that our system works well on more than just Wikipedia text, we also compared our system to $10$ other disambiguation systems over $11$ different datasets.  
This was done by implementing a web-based API for our system that is compatible with GERBIL 1.2.2.~\cite{ref:Usbeck2015}.\footnote{Currently, due to legal restrictions this web-based API is not publically available.}
We refer the interested reader to the GERBIL website\footnote{\url{http://aksw.org/Projects/GERBIL.html}} and paper~\cite{ref:Usbeck2015} for a complete description of these systems and datasets.
The task we considered the \emph{strong annotation task} (D2KB).\footnote{Our system has no spotter, which limits it to strong annotation tasks only.}
In this task we are given an input text containing a number of marked phrases, and must return the entity our system associates with each marked phrase.

In order to produce a fair comparison, we made made two additions to our system for this test.  
The first is a preprocessing step in which we examine the phrase to be annotated and prune it (returning NIL) if it comes from a list of stop words (common verbs).\footnote{We used the list found here: \url{http://www.acme2k.co.uk/acme/3star\%20verbs.htm}.}  
Secondly, if no classifier is found for the phrase to be annotated, we examine the list of DBPedia transitive redirects for the phrase.\footnote{\url{http://wiki.dbpedia.org/Downloads2015-04\#transitive-redirects}}
If the phrase appears as a redirect, we return the corresponding entity associated with the redirect.

We tested our system using all datasets available by default in GERBIL, which are primarily based on news articles, RSS feeds, and tweets.
In Table~\ref{tab:gerbil} we report, for each combination of system and dataset, the micro-F1 (top) and macro-F1 (bottom) scores.
The micro-F1 score is the F1-measure aggregated across annotations, while the macro-F1 score is aggregated across documents.

Firstly, we observe that even though our system does not perform any coherence pruning, it consistently achieves very high macro-F1 scores.
In fact, these macro-F1 scores are the highest in terms of average, $.644$, and lowest in terms of rank-average, $2.45$; Kea comes in second with $.609$ and $2.64$ respectively.
In terms of micro-F1, we fall slightly short of Kea in terms of average and rank-average, $.611$ vs. $.600$ and $2.72$ vs. $3.36$, respectively.

Secondly, our system does extremely well on news.  If we restrict ourselves to the news datasets (ACE2004, AIDA/CoNLL, AQUAINT, MSNBC, N3-Reuters-128, N3-RSS-500), then we achieve the highest average and lowest rank-average scores in terms of both micro-F1 and macro-F1: $.661$/$1.83$ and $.612$/$3$.

Thirdly, our system however performs quite poorly on the KORE50 dataset.  
Many entries in this dataset are single sentences involving very ambiguous entities: since our system does not perform joint disambiguation, these highly ambiguous entities are problematic.
For example, one of the documents in the KORE50 dataset is, ``\underline{David} and \underline{Victoria} added spice to their marriage.'', where the underlined phrases are to be disambiguated.  
The correct entities are ``David Beckham'' and ``Victoria Beckham'', respectively, whereas PML returns the entities ``David'', and ``Victoria Song''.
The first entity is returned due to bias towards in the absense of context, whereas the second result is a performer who appeared in a Korean reality TV show called ``We got Married''.  

We also note that there are systems not currently available to test through GERBIL that nonetheless have published similar comparisons.
One issue however is that these comparisons are made using older versions of GERBIL, and that several changes have been made to recent versions of the system that drastically change these performance numbers.\footnote{See for instance: \url{https://github.com/AKSW/gerbil/issues/98}}
For this reason, we cannot directly compare our system to other recent systems such as that of Ganea et al.~\cite{PBoH2016}, since their published comparison uses a different version of GERBIL.

\section{Full-data Experiment}
\label{sec:full_exp}

In this section, the disambiguation experiment is extended to 
all Wikipedia mentions of more than one candidate senses.
Due to the long processing time of more than 170 million annotations, 
we only run the system with one
setting $(n_{f;w}=100,n_{f;p}=1)$ using multinomial regression.
Also, as in Section \ref{sec:analysis01}, in any dataset, 
if a candidate sense has more than $5000$ labelled annotations,
we sample-without-replacement $5000$ of these annotations 
to reduce the training and validation time. 
The precision results and time statistics are presented in 
Table \ref{tab_analysis04:full_wiki}
and it can be seen that the full performance results are stable and  
comparable to the ones of the corresponding setting in 
Table \ref{tab_analysis01:nword_npart}.

\begin{figure*}[h!tb]
\centering
\begin{adjustbox}{center}
\begin{subfigure}{0.36\textwidth}
    \centering
    \includegraphics[scale=0.58]{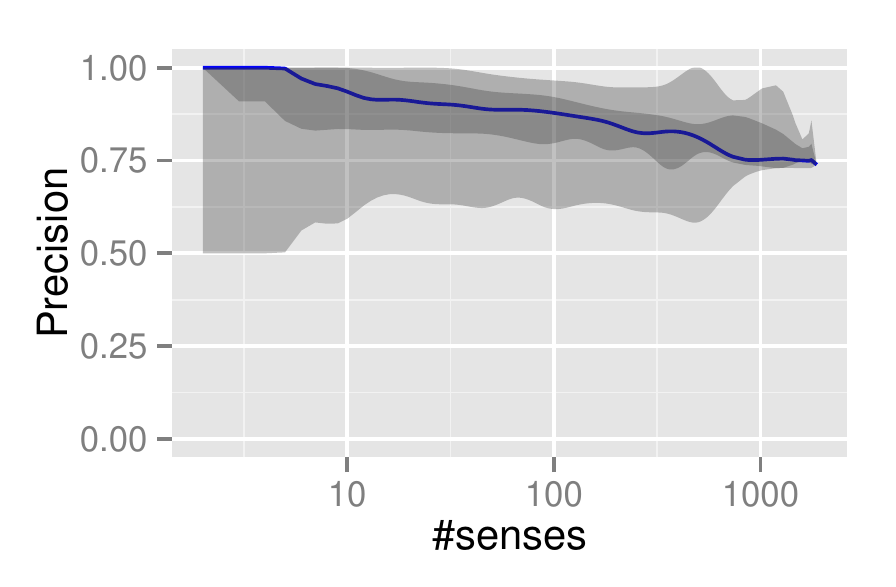}
    \caption{$\mathcal{A}_2$}
    \label{fig_analysis04:precision_trc_a2}
\end{subfigure}
\begin{subfigure}{0.36\textwidth}
    \centering
    \includegraphics[scale=0.58]{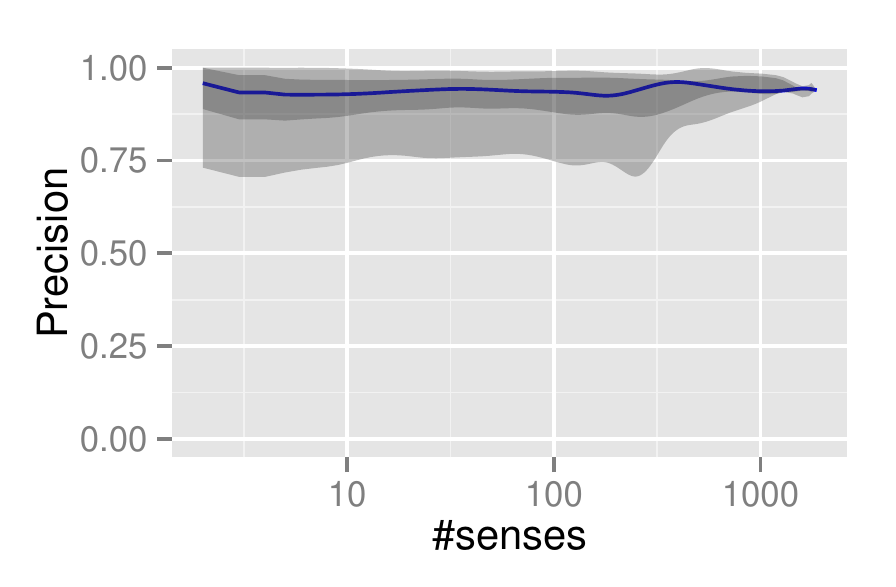}
    \caption{$\mathcal{B}$}
    \label{fig_analysis04:precision_trc_b}
\end{subfigure}
\begin{subfigure}{0.36\textwidth}
    \centering
    \includegraphics[scale=0.58]{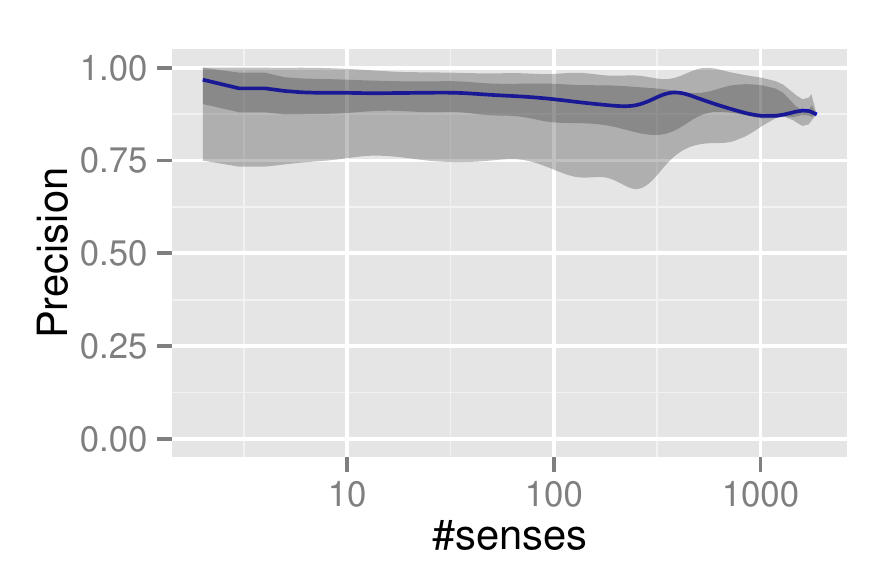}
    \caption{$\mathcal{C}$}
    \label{fig_analysis04:precision_trc_c}
\end{subfigure}
\end{adjustbox}

\begin{subfigure}{0.36\textwidth}
    \centering
    \includegraphics[scale=0.58]{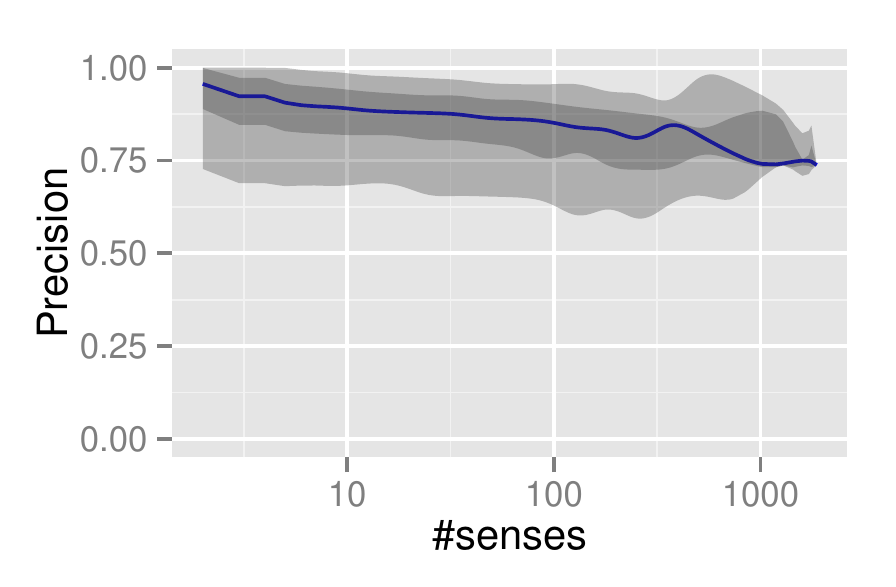}
    \caption{$\mathcal{D}$}
    \label{fig_analysis04:precision_trc_d}
\end{subfigure}
\begin{subfigure}{0.36\textwidth}
    \centering
    \includegraphics[scale=0.58]{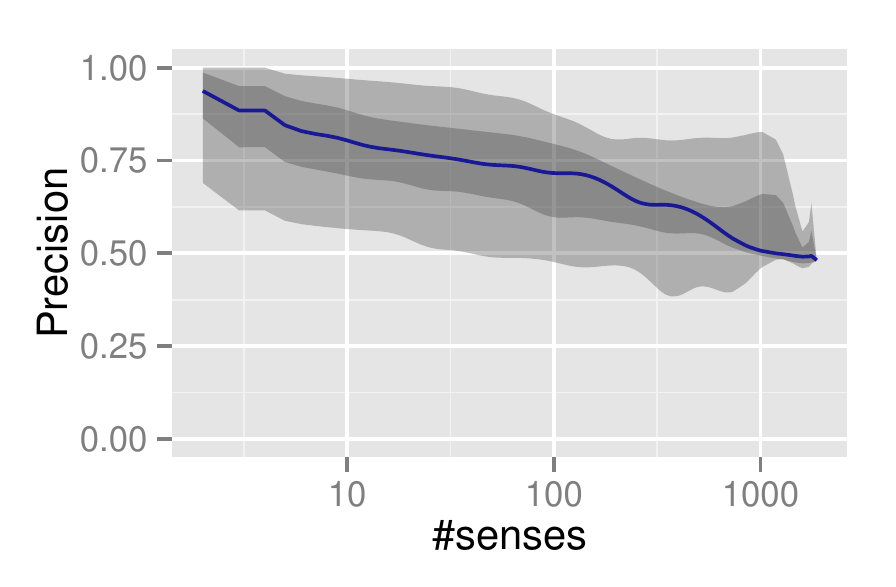}
    \caption{$\mathcal{E}$}
    \label{fig_analysis04:precision_trc_e}
\end{subfigure}
\caption{Per-mention precision for all datasets: the blue line is a smoothed median of precision values and
the two ribbons correspond to the two smoothed quantile intervals $(5\%,95\%)$ and $(25\%,75\%)$}
\label{fig_analysis04:full_wiki}
\end{figure*}

Aside from the global precision, another interesting aspect for this method
is the performance of each unique mention. 
So, in this regard, per-mention precisions for
all mentions are evaluated, smoothed and 
plotted against the number of candidate senses of mentions 
for all datasets
in Figure \ref{fig_analysis04:full_wiki}.
Results for dataset $\mathcal{A}_2$ have wider quantile ranges as
its size is only about $10\%$ of the other datasets'.
As expected, when the number of candidate senses of a mention increases, 
the disambiguation becomes more difficult and the precision reduces.
From Figure \ref{fig_analysis04:precision_trc_c}
to Figure \ref{fig_analysis04:precision_trc_e}, 
the proportional effect of target context size $|a_{context}|$
to per-mention precision can be clearly seen, 
which is especially strong 
with respect to mentions of high number of candidate senses.

\begin{table*}[t]
\caption{Performance results of setting $(n_{f;w}=100,n_{f;p}=1)$ with multinomial regression for the entire Wikipedia}
\centering
\begin{adjustbox}{center}
\begin{tabular}{|c|c|c|c|c|c|c|c|c|c|c|c|}
\hline
\multirow{ 2}{*}{$\mathcal{P}^{\mathcal{A}_2}$}
&\multirow{ 2}{*}{$\underline{\mathcal{P}}^{\mathcal{A}_2}$}
&\multirow{ 2}{*}{$\mathcal{P}^{\mathcal{B}}$}
&\multirow{ 2}{*}{$\underline{\mathcal{P}}^{\mathcal{B}}$}
&\multirow{ 2}{*}{$\mathcal{P}^{\mathcal{C}}$}
&\multirow{ 2}{*}{$\underline{\mathcal{P}}^{\mathcal{C}}$}
&\multirow{ 2}{*}{$\mathcal{P}^{\mathcal{D}}$}
&\multirow{ 2}{*}{$\underline{\mathcal{P}}^{\mathcal{D}}$}
&\multirow{ 2}{*}{$\mathcal{P}^{\mathcal{E}}$}
&\multirow{ 2}{*}{$\underline{\mathcal{P}}^{\mathcal{E}}$}
&$\mathcal{T}_{\text{total}}$ &$\mathcal{T}_{\text{pred}}$\\

&
&&
&&
&&
&&
&$(\times10^3 \text{s})$ &$(\text{ms})$\\
\hline
.9188 &.9220 &.9261 &.9172 &.9238 &.9265 &.9012 &.9067 &.8617 &.8712 &1400.77 &2.82\\
\hline
\end{tabular}
\end{adjustbox}
\label{tab_analysis04:full_wiki}

\end{table*}

\section{Conclusions}
\label{sec:conclusion}

This paper proposes a new per-mention learning (PML) disambiguation system, in which the feature engineering and model training is done per unique mention. 
The most significant  advantage of this approach lies in the specialized learning that improves the precision. 
The method is also highly parallelizable and scalable, and can be further tuned for higher precision with per-mention or per-candidate pruning. 
Furthermore, this per-mention disambiguation approach can be easily calibrated or tuned for specific mentions with new datasets, without affecting the results of other mentions.
In an extensive comparisons with $10$ other disambiguation systems over $11$ different datasets, we have shown that our PML system is very competitive with the state-of-the-art, and, for the specific case ofdisambiguating news, consistently outperforms these other systems.

There are several potential key tasks that we consider for future work. 
Firstly, we wish to combine PML with graph-based approaches for further improvement. 
One method is to impose coherence between chosen senses in the same document or sentence, by using a fast graph-based approach on the top-$k$ results of our proposed disambiguation. 
Secondly, the computational bottle neck of our system for new annotations lies in the set matching operation between annotation and sense context, which can be improved with an optimized algorithm.


\bibliographystyle{abbrv}
\bibliography{disambiguation_arxiv} 

\begin{thebibliography}{10}

\bibitem{ref:Carmel2014}
D.~Carmel, M.-W. Chang, E.~Gabrilovich, B.-J.~P. Hsu, and K.~Wang.
\newblock {ERD'14: Entity recognition and disambiguation challenge}.
\newblock {\em SIGIR Forum}, 48(2):63--77, Dec. 2014.

\bibitem{ref:Cucerzan2007}
S.~Cucerzan.
\newblock Large-scale named entity disambiguation based on wikipedia data.
\newblock In {\em Proceedings of EMNLP-CoNLL 2007}, pages 708--716, June 2007.

\bibitem{ref:Cucerzan2014}
S.~Cucerzan.
\newblock {Name entities made obvious: the participation in the ERD 2014
  evaluation}.
\newblock In {\em {Proceedings of the First International Workshop on Entity
  Recognition \& Disambiguation}}, ERD, pages 95--100, New York, NY, USA, 2014.
  ACM.

\bibitem{ref:Daiber2013}
J.~Daiber, M.~Jakob, C.~Hokamp, and P.~N. Mendes.
\newblock Improving efficiency and accuracy in multilingual entity extraction.
\newblock In {\em Proceedings of the 9th International Conference on Semantic
  Systems}, I-SEMANTICS, 2013.

\bibitem{ref:Ferragina2010A}
P.~{Ferragina} and U.~{Scaiella}.
\newblock {Fast and accurate annotation of short texts with Wikipedia pages}.
\newblock {\em ArXiv e-prints}, June 2010.

\bibitem{ref:Ferragina2010B}
P.~Ferragina and U.~Scaiella.
\newblock {TAGME: On-the-fly annotation of short text fragments (by Wikipedia
  entities)}.
\newblock In {\em Proceedings of the 19th ACM International Conference on
  Information and Knowledge Management}, CIKM, pages 1625--1628, 2010.

\bibitem{ref:Ferrucci2012}
D.~A. Ferrucci.
\newblock {Introduction to "This is Watson"}.
\newblock {\em IBM Journal of Research and Development}, 56(3):235--249, May
  2012.

\bibitem{PBoH2016}
O.~Ganea, M.~Ganea, A.~Lucchi, C.~Eickhoff, and T.~Hofmann.
\newblock Probabilistic bag-of-hyperlinks model for entity linking.
\newblock In {\em Proceedings of the 25th International Conference on World
  Wide Web, {WWW} 2016, Montreal, Canada, April 11 - 15, 2016}, pages 927--938,
  2016.

\bibitem{GuoB14}
Z.~Guo and D.~Barbosa.
\newblock Robust entity linking via random walks.
\newblock In {\em Proceedings of the 23rd {ACM} International Conference on
  Conference on Information and Knowledge Management, {CIKM} 2014, Shanghai,
  China, November 3-7, 2014}, pages 499--508, 2014.

\bibitem{ref:Han2011A}
X.~Han and L.~Sun.
\newblock A generative entity-mention model for linking entities with knowledge
  base.
\newblock In {\em Proceedings of the 49th Annual Meeting of the Association for
  Computational Linguistics: Human Language Technologies}, HLT, pages 945--954,
  2011.

\bibitem{ref:Han2011B}
X.~Han, L.~Sun, and J.~Zhao.
\newblock {Collective entity linking in web text: A graph-based method}.
\newblock In {\em Proceedings of the 34th International ACM SIGIR Conference on
  Research and Development in Information Retrieval}, SIGIR, pages 765--774,
  2011.

\bibitem{ref:Hoffart2013}
J.~Hoffart.
\newblock Discovering and disambiguating named entities in text.
\newblock In {\em Proceedings of the 2013 SIGMOD/PODS Ph.D. Symposium}, SIGMOD
  PhD Symposium, pages 43--48, 2013.

\bibitem{HulpusPH15}
I.~Hulpus, N.~Prangnawarat, and C.~Hayes.
\newblock Path-based semantic relatedness on linked data and its use to word
  and entity disambiguation.
\newblock In {\em Proceedings of the 14th International Semantic Web Conference
  {ISWC}}, pages 442--457, 2015.

\bibitem{ref:Kulkarni2009}
S.~Kulkarni, A.~Singh, G.~Ramakrishnan, and S.~Chakrabarti.
\newblock {Collective annotation of Wikipedia entities in web text}.
\newblock In {\em Proceedings of the 15th ACM SIGKDD International Conference
  on Knowledge Discovery and Data Mining}, KDD, pages 457--466, 2009.

\bibitem{ref:McNamee2010}
P.~McNamee.
\newblock {HLTCOE} efforts in entity linking at {TAC KBP} 2010.
\newblock In {\em Proceedings of the TAC 2010 Workshop}, 2010.

\bibitem{ref:Meij2012}
E.~Meij, W.~Weerkamp, and M.~de~Rijke.
\newblock Adding semantics to microblog posts.
\newblock In {\em Proceedings of the Fifth International Conference on Web
  Search and Web Data Mining}, WSDM, pages 563--572, 2012.

\bibitem{ref:Milne2008}
D.~Milne and I.~H. Witten.
\newblock {Learning to link with Wikipedia}.
\newblock In {\em Proceedings of the 17th ACM Conference on Information and
  Knowledge Management}, CIKM, pages 509--518, 2008.

\bibitem{ref:Moro2014}
A.~Moro, A.~Raganato, and R.~Navigli.
\newblock Entity linking meets word sense disambiguation: a unified approach.
\newblock {\em Transactions of the Association for Computational Linguistics},
  2:231--244, 2014.

\bibitem{ref:Olieman2014}
A.~Olieman, H.~Azarbonyad, M.~Dehghani, J.~Kamps, and M.~Marx.
\newblock {Entity linking by focusing DBpedia candidate entities}.
\newblock In {\em {Proceedings of the First International Workshop on Entity
  Recognition \& Disambiguation}}, ERD, pages 13--24, 2014.

\bibitem{ref:Piccinno2014}
F.~Piccinno and P.~Ferragina.
\newblock {From TAGME to WAT: A new entity annotator}.
\newblock In {\em {Proceedings of the First International Workshop on Entity
  Recognition \& Disambiguation}}, ERD, pages 55--62, 2014.

\bibitem{ref:Qureshi2014}
M.~A. Qureshi, C.~O'Riordan, and G.~Pasi.
\newblock Exploiting wikipedia for entity name disambiguation in tweets.
\newblock In {\em Natural Language Processing and Information Systems - 19th
  International Conference on Applications of Natural Language to Information
  Systems, {NLDB}}, pages 184--195, 2014.

\bibitem{ref:Suchanek2013}
F.~Suchanek and G.~Weikum.
\newblock {Knowledge harvesting in the big-data era}.
\newblock In {\em Proceedings of the 2013 ACM SIGMOD International Conference
  on Management of Data}, SIGMOD, pages 933--938, New York, NY, USA, 2013. ACM.

\bibitem{AGDISTIS2014}
R.~Usbeck, A.~N. Ngomo, M.~R{\"{o}}der, D.~Gerber, S.~A. Coelho, S.~Auer, and
  A.~Both.
\newblock {AGDISTIS} - agnostic disambiguation of named entities using linked
  open data.
\newblock In {\em Proceedings of the 21st European Conference on Artificial
  Intelligence {ECAI}}, pages 1113--1114, 2014.

\bibitem{ref:Usbeck2015}
R.~Usbeck, M.~R\"{o}der, A.-C. Ngonga~Ngomo, C.~Baron, A.~Both, M.~Br\"{u}mmer,
  D.~Ceccarelli, M.~Cornolti, D.~Cherix, B.~Eickmann, P.~Ferragina, C.~Lemke,
  A.~Moro, R.~Navigli, F.~Piccinno, G.~Rizzo, H.~Sack, R.~Speck, R.~Troncy,
  J.~Waitelonis, and L.~Wesemann.
\newblock {GERBIL: general entity annotator benchmarking framework}.
\newblock In {\em Proceedings of the 24th International Conference on World
  Wide Web}, WWW, pages 1133--1143, 2015.

\end{thebibliography}

\end{document}